\documentclass{article}

\usepackage{amsmath,amsfonts,bm}

\def\eqref#1{equation~\ref{#1}}

\def\1{\bm{1}}

\def\vw{{\bm{w}}}

\def\mK{{\bm{K}}}

\def\mP{{\bm{P}}}

\def\mX{{\bm{X}}}

\DeclareMathAlphabet{\mathsfit}{\encodingdefault}{\sfdefault}{m}{sl}
\SetMathAlphabet{\mathsfit}{bold}{\encodingdefault}{\sfdefault}{bx}{n}

\newcommand{\R}{\mathbb{R}}

\PassOptionsToPackage{numbers, compress}{natbib}

     \usepackage[preprint]{neurips_2019}

\usepackage[utf8]{inputenc} 
\usepackage[T1]{fontenc}    
\usepackage{hyperref}       
\usepackage{url}            
\usepackage{booktabs}       
\usepackage{amsfonts}       
\usepackage{nicefrac}       
\usepackage{microtype}      
\usepackage[titlenumbered,ruled]{algorithm2e}
\usepackage{algorithmic}
\usepackage{multirow}
\usepackage{multicol}
\usepackage{makecell}
\usepackage{amsmath}
\usepackage{graphicx}
\usepackage{caption}
\usepackage{subcaption}
\newcommand{\mr}[2]{\multirow{#1}{*}{\begin{tabular}[c]{@{}c@{}}#2\end{tabular}}}

\title{Learning Low-Rank Approximation for CNNs}

\author{
Dongsoo Lee\textsuperscript{1}, Se Jung Kwon\textsuperscript{1}, Byeongwook Kim\textsuperscript{1} and Gu-Yeon Wei\textsuperscript{1,2} \\
  \textsuperscript{1} Samsung Research, Seoul, Republic of Korea \\
  \textsuperscript{2} Harvard University, Cambridge, MA \\
  \texttt{\{dongsoo3.lee, sejung0.kwon, byeonguk.kim, gy.wei\}@samsung.com}\\
}

\begin{document}

\maketitle

\begin{abstract}
  Low-rank approximation is an effective model compression technique to not only reduce parameter storage requirements, but to also reduce computations.
  For convolutional neural networks (CNNs), however, well-known low-rank approximation methods, such as Tucker or CP decomposition, result in degraded model accuracy because decomposed layers hinder training convergence.
  In this paper, we propose a new training technique that finds a flat minimum in the view of low-rank approximation without a decomposed structure during training.
  By preserving the original model structure, 2-dimensional low-rank approximation demanding lowering (such as im2col) is available in our proposed scheme.
  We show that CNN models can be compressed by low-rank approximation with much higher compression ratio than conventional training methods while maintaining or even enhancing model accuracy.
  We also discuss various 2-dimensional low-rank approximation techniques for CNNs.
  
\end{abstract}

\section{Introduction}

Deep neural networks (DNNs) are usually over-parameterized to expedite the local minimum search process \cite{lottery, deeplearningbook, SHan_2015}.
Hence, various model compression techniques have been proposed to reduce memory footprint and the amount of computations required for inference \cite{xu2018alternating, suyog_prune, distillation2018}.
For example, parameter pruning enables DNNs to be sparse by zeroing out many parameter values.
Parameter quantization uses fewer bits while maintaining comparable model accuracy.
Parameter pruning and quantization, however, demand dedicated hardware designs to support sparsity and bit-level manipulation in memory and/or computation units in order to maximize model reduction benefits \cite{EIE, lee2018viterbibased, viterbi_quantized}.
Low-rank approximation, on the other hand, does not require specialized hardware. Rather, model structure is simplified by reducing parameter count.

Low-rank approximation has been successfully applied to speech recognition \cite{SVD_Projection} and language models \cite{GroupReduce} where DNNs employ 2-dimensional (2D) weight matrices for long short-term memory (LSTM) or fully-connected layers.
In the case of 2D low-rank approximation, singular-value decomposition (SVD) minimizes the Frobenius norm of the difference between the original matrix and the approximated matrix.
Yet, SVD cannot be directly utilized for convolutions in CNNs because weights need to be represented by higher-dimensional (e.g., 4D) tensors.
Thus, Tucker decomposition \cite{tucker_samsung} or CP decomposition \cite{cp_decomposition}, which is applicable to tensors of any dimensions, is appropriate low-rank approximation methods for CNNs.

Tucker decomposition or CP decomposition decomposes a tensor into multiple tensors to be multiplied in a consecutive manner.
If training is performed on the transformed network structure with consecutive tensors without activation functions between them, then convergence can be degraded due to vanishing or exploding gradients \cite{deeplearningbook}.
Indeed, drop in model accuracy is observed for CNNs decomposed by Tucker or CP decomposition even after model retraining as a fine-tuning process \cite{cp_decomposition, tucker_samsung}.
Note that dimensionality reduction techniques (e.g., im2col \cite{tiling_matmul}) can transform weight tensors to 2D matrices.
For such a case, however, fine-tuning after pre-training is not available since lowering cannot preserve the transformed model during retraining.

To address the aforementioned limitation, this paper introduces a new training algorithm, called DeepTwist, that improves the quality of low-rank approximations for CNNs.
In the proposed method, we retain the original model structure without considering low-rank approximation.
We show that occasional weight value distortions, based on low-rank approximation, significantly improves overall model accuracy compared to conventional training.
Moreover, by obviating the need to maintain a decomposed structure after low-rank approximation, our proposed method can combine lowering methods with SVD.
We discuss various SVD-based low-rank approximation techniques for CNNs to enhance compression ratio even further compared with Tucker decomposition.

\section{Training Algorithm for Low-Rank Approximation}

\begin{figure}[t]
\begin{center}
\includegraphics[width=1.0\linewidth]{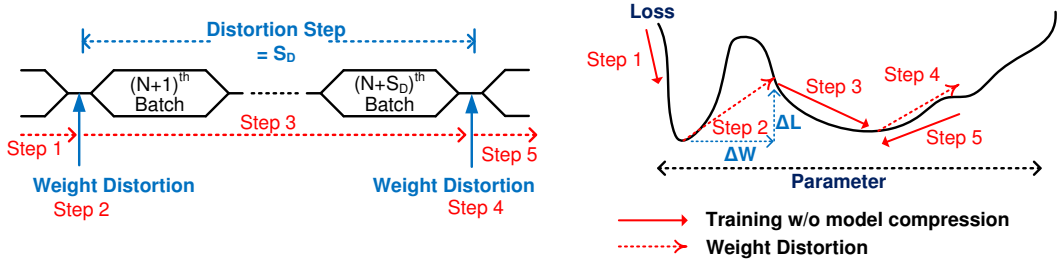}
\caption{DeepTwist optimization procedure and corresponding local minimum search process. A model is trained without considering low-rank approximation for each batch. Weights of a model, however, are distorted by low-rank approximation at every $S_D$ distortion steps.}
\label{fig:deeptwist_illustration}
\end{center}
\end{figure}

Each parameter inherits a certain amount of error from low-rank approximation.
SVD, CP decomposition, and Tucker decomposition were designed to reduce approximation error given a rank or a set of ranks.
In contrast, the underlying principle of our proposed training algorithm is to find a particular local minimum of flat loss surface, robust to parameter errors induced by low-rank approximation.
In the context of model compression, flatness is measured by the loss function increase when the model is compressed right after reaching a local minimum ($\Delta \mathcal{L}$ in Figure \ref{fig:deeptwist_illustration}).
Note that local minima with similar model accuracy can exhibit significantly different accuracy drops after low-rank approximation depending on the smoothness of the local loss surface \cite{DeepTwist, quant_lee}.
Unlike conventional retraining methods (for low-rank approximation) to be performed after transforming a model, our proposed scheme preserves the original model structure and searches for a local minimum well suited for low-rank approximation.

Our training procedure for low-rank approximation, called DeepTwist, is illustrated in Figure \ref{fig:deeptwist_illustration}.
At each weight distortion step (after training every $S_D$ batches), parameter tensors are decomposed into multiple tensors following low-rank approximation.
Then, those decomposed tensors are multiplied to reconstruct the original tensor form.
In other words, we effectively inject noise into the parameter tensors without modifying the structure.
Note that the training procedure is always completed at a weight distortion step to permit a transformed structure to be deployed for inference.
Other than weight distortion steps that evaluate flatness of a loss surface occasionally, low-rank approximation is not considered while training the model.
If a particular local minimum is not flat enough, against the amount of weight error produced by low-rank approximation, our training procedure escapes such a local minimum through a weight distortion step (such an example is illustrated as Step 2 in Figure \ref{fig:deeptwist_illustration}).
Otherwise, the optimization process is continued in the search space around a local minimum (for instance, Steps 4 and 5 in Figure \ref{fig:deeptwist_illustration}).

Determining the best distortion step $S_D$ is important.
Given a parameter set $\vw$ and a learning rate $\gamma$, the loss function of a model $\mathcal{L}(\vw)$ can be approximated as 
\begin{equation}
    \label{eq:eq_loss_approxi}
    \mathcal{L}(\vw) \simeq \mathcal{L}(\vw_0) + (\vw - \vw_0)^\top (H(\vw_0)/2) (\vw - \vw_0) 
\end{equation}
using a local quadratic approximation where $H$ is the Hessian of $\mathcal{L}$ and $\vw_0$ is a set of parameters at a local minimum.
Then, $\vw$ can be updated by gradient descent as follows:
\begin{equation}
    \label{eq:eq_1}
    \vw_{t+1} = \vw_{t} - \gamma \frac{\partial \mathcal{L}}{\partial \vw}\big|_{\vw=\vw_t}  \simeq \vw_{t} - \gamma H(\vw_0)(\vw_t - \vw_0).
\end{equation}
Thus, after $S_D$ steps, we obtain $\vw_{t+S_D}$ = $\vw_0$ + $(I-\gamma H(\vw_0))^{S_D}$ $(\vw_t - \vw_0)$, where $I$ is an identity matrix.
Suppose that $H$ is positive semi-definite and all elements of $I - \gamma H(\vw_0)$ are less than 1.0, large $S_D$ allows $\vw_{t+S_D}$ to converge to $\vw_0$.
Correspondingly, large $S_D$ is usually preferred for our proposed training method, DeepTwist, to converge well.
Large $S_D$ is also desirable to avoid stochastic variance (from the batch selection) and to reduce computational overhead from weight distortion steps (where decomposition is performed).
Too large of a $S_D$, however, yields fewer opportunities to escape local minima especially when the learning rate has decayed.
In practice, $S_D$ falls in the range of 100 to 5000 steps.

Controlling learning rate has been introduced as a technique to explore various local minima.
For example, a nonmonotonic scheduling of learning rates is proposed in \cite{kaist_learning} where learning rates can increase, aimed at reaching a flatter minimum.
Similarly, cyclical learning rates have also been explored \cite{cyclical_learning}.
Another approach to avoid sharp minima is to apply weight noise to parameters as a regularization technique \cite{flat_minima}.
Note that DeepTwist is distinguished from those previous attempts because 1) noise is added only when a set of parameters is close to a local minimum to avoid unnecessary escapes, and 2) escape distance (=$\Delta \vw$ in Figure~\ref{fig:deeptwist_illustration}) is determined by noise induced by low-rank approximation (thus, DeepTwist is a compression-aware training algorithm).

\section{Tucker Decomposition Optimized by DeepTwist}

In this section, we apply our proposed DeepTwist training algorithm integrated with Tucker decomposition \cite{tucker} to CNN models and demonstrate supriority of DeepTwist over conventional training methods.
In CNNs, the convolution operation requires a 4D kernel tensor $\mathcal{K}=$ $\mathbb{R}^{d \times d \times S \times T}$ where each kernel has $d\times d$ dimension, $S$ is the input feature map size, and $T$ is the output feature map size.
Then, following the Tucker decomposition algorithm, $\mathcal{K}$ is decomposed into three components as
\begin{equation}
    \label{eq:eq_tucker}
    \mathcal{\Tilde{K}}_{i,j,s,t} = \sum_{r_s=1}^{R_s}\sum_{r_t=1}^{R_t}\mathcal{C}_{i,j,r_s,r_t}\mP_{s,r_s}^S\mP_{t,r_t}^T ,
\end{equation}
where $\mathcal{C}_{i,j,r_s,r_t}$ is the reduced kernel tensor, $R_s$ is the rank for input feature map dimension, $R_t$ is the rank for output feature map dimension, and $\mP^S$ and $\mP^T$ are 2D filter matrices to map $\mathcal{C}_{i,j,r_s,r_t}$ to $\mathcal{K}_{i,j,s,t}$.
Each component is obtained to minimize the Frobenius norm of ($\mathcal{\Tilde{K}}_{i,j,s,t}$ $-$ $\mathcal{K}_{i,j,s,t}$).
As a result, one convolution layer is divided into three convolution layers, specifically, $(1 \times 1)$ convolution for $\mP^S$, $(d \times d)$ convolution for $\mathcal{C}_{i,j,r_s,r_t}$, and $(1 \times 1)$ convolution for $\mP^T$ \cite{tucker_samsung}.

In prior tensor decomposition schemes, model training is performed as a fine-tuning procedure after the model is restructured and fixed \cite{cp_decomposition, tucker_samsung}.
On the other hand, DeepTwist training algorithm in Figure \ref{fig:deeptwist_illustration} is conducted for Tucker decomposition as follows:
\begin{itemize}
\item [Step 1:] Perform normal training for $S_D$ steps (batches) without considering Tucker decomposition
\item [Step 2:] Calculate $\mathcal{C}$, $\mP^S$, and $\mP^T$ using Tucker decomposition to obtain $\mathcal{\Tilde{K}}$
\item [Step 3:] Replace $\mathcal{K}$ with $\mathcal{\Tilde{K}}$ (a weight distortion step in Figure~\ref{fig:deeptwist_illustration})
\item [Step 4:] Go to Step 1 with updated $\mathcal{K}$
\end{itemize}
After repeating a number of the above steps towards convergence, the entire training process should stop at Step 2, and then the final decomposed structure is extracted for inference.
Because the model is not restructured except in the last step, Steps 2 and 3 can be regarded as special steps to encourage a flat local minimum where weight noise by decomposition does not noticeably degrade the loss function.

Using the pre-trained ResNet-32\footnote{https://github.com/akamaster/pytorch\_resnet\_cifar10}  model with CIFAR-10 dataset \cite{resnet, tensorly}, we compare two training methods for Tucker decomposition: 1) typical training with a decomposed model and 2) DeepTwist training, which maintains the original model structure and occasionally injects weight noise through decomposition.
Using an SGD optimizer, both training methods follow the same learning schedule: learning rate is 0.1 for the first 100 epochs, 0.01 for the next 50 epochs, and 0.001 for the last 50 epochs.
Except for the first layer, which is much smaller than the other layers, all convolution layers are compressed by Tucker decomposition with rank $R_s$ and $R_t$ selected to be $S$ and $T$ multiplied by a constant number $R_c$ ($0.3\le R_c\le0.7$ in this experiment).
Then, the compression ratio of a convolution layer is $d^2ST/(SR_s+d^2R_sR_t+TR_t)$ $=d^2ST/(S^2R_c+d^2R_c^2ST+T^2R_c)$, which can be approximated to be $1/R^2_c$ if $S=T$ and $d\gg R_c$.
For DeepTwist, $S_D$ is chosen to be 200.


\begin{figure}[t]
    \centering
    \includegraphics[width=0.6\textwidth]{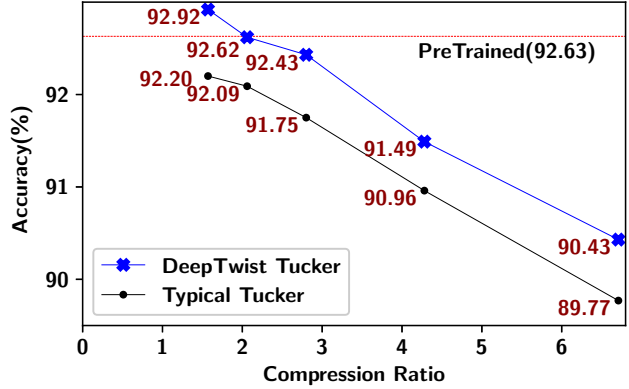}
    \caption{Test accuracy comparison on ResNet-32 using CIFAR-10 trained by typical training method and DeepTwist method with various compression ratios. For DeepTwist, test accuracy is measured only at Step 3 that allows to extract a decomposed structure, and $S_D$ is 200.}
    \label{fig:tucker1}
\end{figure}

\begin{figure}[t]
    \centering
    \includegraphics[width=1.0\textwidth]{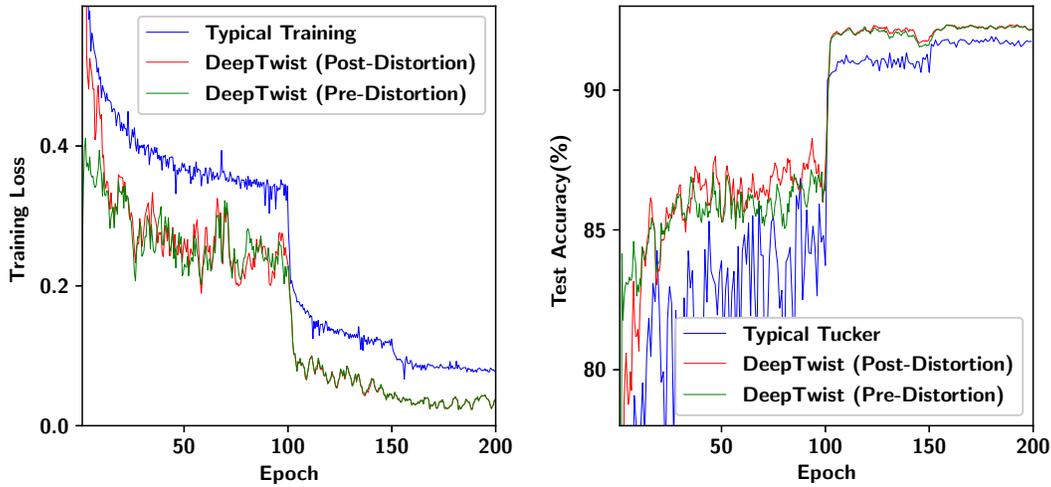}
    \caption{Training loss and test accuracy of ResNet-32 using CIFAR-10. For DeepTwist, training loss and test accuracy are only monitored right before or after a distortion step (pre-distortion or post-distortion). Compression ratio is 2.8 with $R_c$=0.5.}
    \label{fig:tucker2}
\end{figure}

Figure~\ref{fig:tucker1} shows test accuracy after Tucker decomposition\footnote{https://github.com/larry0123du/Decompose-CNN} by two different training methods.
Note that throughout this paper, all test accuracy results using DeepTwist are evaluated only at Step 3 where the training process can stop to generate a decomposed structure.
In Figure~\ref{fig:tucker1}, across a wide range of compression ratios (determined by $R_c$), DeepTwist yields higher model accuracy compared to typical training.
Note that even higher model accuracy than that of the pre-trained model can be achieved by DeepTwist if the compression ratio is small enough.
In fact, Figure~\ref{fig:tucker2} shows DeepTwist improves training loss and test accuracy throughout the entire training process.
Initially, the gap of training loss and test accuracy between pre-distortion and post-distortion is large.
Such a gap, however, is quickly reduced through training epochs, because a local minimum found by DeepTwist exhibits a flat loss surface in view of low-rank approximation.
Overall, ResNet-32 converges successfully through the entire training process with lower training loss and higher test accuracy compared with a typical training method.

\begin{figure}[t]
\begin{center}
\includegraphics[width=0.9\linewidth]{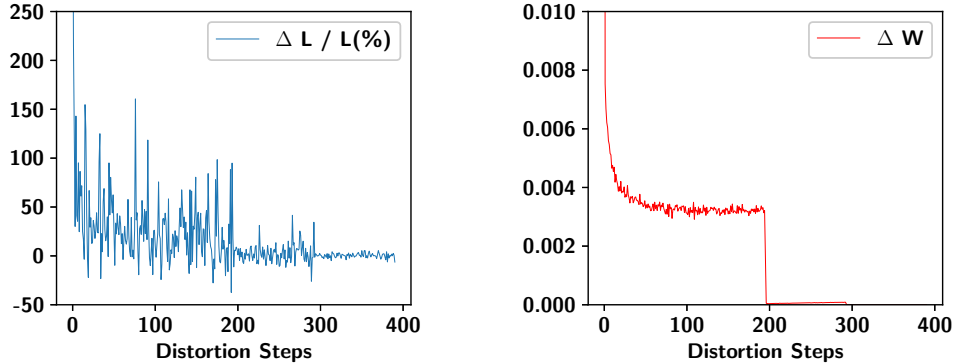}
\caption{Difference of training loss function and average Frobenius norm of weight values by Step 2 and Step 3 of DeepTwist. $R_c$ $=0.5$ and $S_D$ $=200$ are used.}
\label{fig:delta_values}
\end{center}
\end{figure}

To investigate its effect on local minima exploration, Figure~\ref{fig:delta_values} presents the changes of loss function and weight magnitude values incurred by DeepTwist.
In Figure~\ref{fig:delta_values}(left), $\Delta \mathcal{L}/\mathcal{L}$ is given as the loss function increase $\Delta\mathcal{L}$ (due to a weight distortion step) divided by $\mathcal{L}$, which is the loss function value right before a weight distortion step.
In Figure~\ref{fig:delta_values}(right), $\Delta \vw$ is defined as $||\vw - \Tilde{\vw}||^2_{\mathcal{F}}$ $/N(\vw)$, where $\vw$ is the entire set of weights to be compressed, $\Tilde{\vw}$ is the set of weights distorted by Tucker decomposition through Step 2 of DeepTwist, $N(\vw)$ is the number of elements of $\vw$, and $||\mX||^2_{\mathcal{F}}$ is the Frobenius norm of $\mX$.
Initially, $\vw$ fluctuates with large corresponding $\Delta \mathcal{L}$.
Then, both $\Delta \mathcal{L}$ and $\Delta \vw$ decrease and  Figure~\ref{fig:delta_values} shows that DeepTwist finds flatter local minima (in the view of Tucker decomposition) successfully.
When the learning rate is reduced at 100th and 150th epochs (roughly corresponding to the 200th and 300th distortion steps), $\Delta \mathcal{L}$ and $\Delta \vw$ decrease significantly because of a lot reduced local minima exploration space.
In other words, DeepTwist helps an optimizer to detect a local minimum where Tucker decomposition does not alter the loss function value noticeably.

\section{2-Dimensional SVD Enabled by DeepTwist}

In this section, we discuss why 2D SVD needs to be investigated for CNNs and how DeepTwist enables a training process for 2D SVD.

\subsection{Issues of 2D SVD on Convolution Layers}

Convolution can be performed by matrix multiplication if an input matrix is transformed into a Toeplitz matrix with redundancy and a weight kernel is reshaped into a $T \times (S \times d \times d)$ matrix (i.e., a lowered matrix) \cite{lowering}.
Then, commodity computing systems (such as CPUs and GPUs) can use libraries such as Basic Linear Algebra Subroutines (BLAS) without dedicated hardware resources for convolution \cite{minsik_mec}.
Some recently developed DNN accelerators, such as Google's Tensor Processing Unit (TPU) \cite{TPU}, are also focused on matrix multiplication acceleration (usually with reduced precision).

For BLAS-based CNN inference, reshaping a 4D tensor $\mathcal{K}$ and performing SVD is preferred for low-rank approximation rather than relatively inefficient Tucker decomposition followed by a lowering technique. 
However, a critical problem with SVD (with a lowered matrix) for convolution layers is that two decomposed matrices by SVD do not present corresponding (decomposed) convolution layers, because of intermediate lowering steps.
As a result, fine-tuning methods requiring a structurally modified model for training are not available for convolution layers to be compressed by SVD.
On the other hand, DeepTwist does not alter the model structure for training.
For DeepTwist, SVD can be performed as a way to feed noise into a weight kernel $\mathcal{K}$ for every distortion step.
Once DeepTwist stops at a distortion step, the final weight values can be decomposed by SVD and used for inference with reduced memory footprint and computations.
In other words, DeepTwist enables SVD-aware training for CNNs.

\begin{figure*}[t]
\hfill
	\begin{minipage}[t]{.54\textwidth}
		\includegraphics[width=1\linewidth]{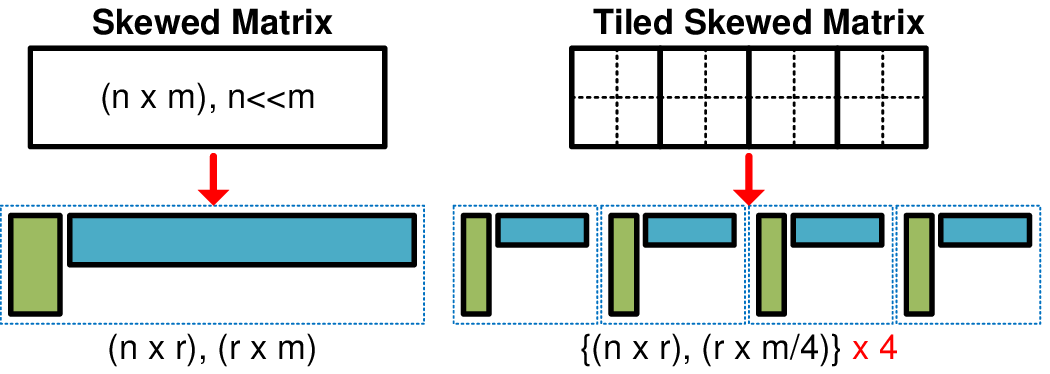}
	\end{minipage}
	\hfill
	\begin{minipage}[t]{.44\textwidth}
		\includegraphics[width=1\linewidth]{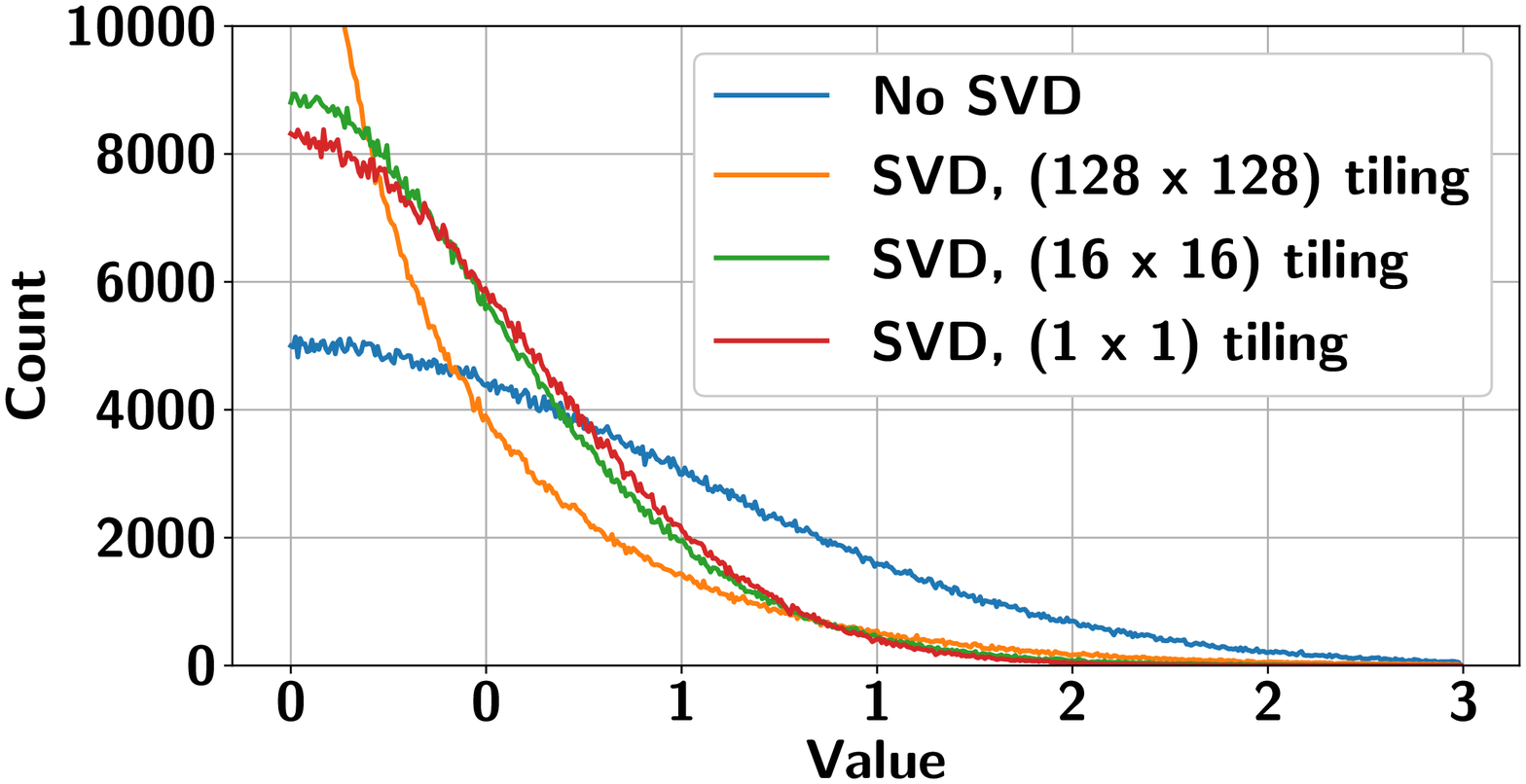}
	\end{minipage}
	\hfill
	\caption{Skewed matrix and a tiling technique are illustrated on the left side, while the right side presents distributions of weights after SVD with different tiling schemes (only positive weights are included).}
	\label{fig:nmf_tiling2}
\end{figure*}

\subsection{Tiling-Based SVD for Skewed Weight Matrices}

A reshaped kernel matrix $\mK$ $\in\R^{T\times (S \times d \times d)}$ is usually a skewed matrix where row-wise dimension ($n$) is smaller than column-wise dimension ($m$) as shown in Figure~\ref{fig:nmf_tiling2} (i.e., $n \ll m$).
A range of available rank $r$ for SVD, then, is constrained by small $n$ and the compression ratio is approximated to be $n/r$.
If such a skewed matrix is divided into four tiles as shown in Figure~\ref{fig:nmf_tiling2} and the four tiles do not share much common chateracteristics, then tiling-based SVD can be a better approximator and rank $r$ can be further reduced without increasing approximation error.
Moreover, fast matrix multiplication is usually implemented by a tiling technique in hardware to improve the weight reuse rate \cite{tiling_matmul}.
Hence, tiling could be a natural choice not only for high-quality SVD but also for high-performance hardware operations.

To investigate the impact of tiling on weight distributions after SVD, we tested a $(1024\times1024)$ random weight matrix in which elements follow a Gaussian distribution.
A weight matrix is divided by $(1\times1)$, $(16\times16)$, or $(128\times128)$ tiles (then, each tile is a submatrix of $(1024\times1024)$, $(64\times64)$, or $(8\times8)$ size).
Each tile is compressed by SVD to achieve the same overall compression ratio of $4\times$ for all of the three cases.
As described in Figure~\ref{fig:nmf_tiling2} (on the right side), increasing the number of tiles tends to increase the count of near-zero and large weights (i.e., variance of weight values increases).
Figure~\ref{fig:nmf_tiling2} can be explained by sampling theory where decreasing the number of random samples (of small tile size) increases the variance of sample mean.
In short, tiling affects the variance of weights after SVD (while the impact of such variance on model accuracy should be empirically studied).

\begin{table}[]
\begin{center}
\caption{Test accuracy(\%) of ResNet-32 model using CIFAR-10 dataset while the 9 largest convolution layers ($T$=$S$=64, $d$=3) are compressed by SVD using different tiling configurations. For each tile size, rank $r$ is selected to achieve compression ratio of $2\times$ or $4\times$. $S_D$=200 is used for DeepTwist.} 
\label{table:tileacc}
\begin{tabular}{c||c|ccccc}
\Xhline{2\arrayrulewidth}
            \mr{2}{Pre-Trained} & Compression & \multicolumn{4}{c}{\begin{tabular}[c]{@{}c@{}}Size of Each Tile\end{tabular}} \\
            \cline{3-6}
            & Ratio & 64$\times$64 & 32$\times$32 & 16$\times$16 & 8$\times$8   \\
\Xhline{2\arrayrulewidth}
 \mr{2}{92.63} & 2$\times$    & 93.34 ($r$=16) & 93.11 ($r$=8) & 93.01 ($r$=4) & 93.23 ($r$=2) \\
 & 4$\times$   & 92.94 ($r$=8)~~ & 92.97 ($r$=4) & 93.00 ($r$=2) & 92.81 ($r$=1) \\
\Xhline{2\arrayrulewidth}
\end{tabular}
\end{center}
\end{table}

We applied the tiling technique and SVD to the 9 largest convolution layers of ResNet-32 using the CIFAR-10 dataset.
Weights of selected layers are reshaped into $64\times(64\times3\times3)$ matrices with the tiling configurations described in Table~\ref{table:tileacc}.
DeepTwist performs training with the same learning schedule and $S_D$(=200) used in Section 3.
Compared to the test accuracy of the pre-trained model (=92.63\%), all of the compressed models in Table~\ref{table:tileacc} achieves higher model accuracy due to the regularization effect of DeepTwist.
Note that for each target compression ratio, the relationship between tile size and model accuracy is not clear.
Hence, various configurations of tile size need to be explored to enhance model accuracy, even though variation of model accuracy for different tile size is small.


\begin{figure}[t]
\begin{center}
    \centering
    \includegraphics[width=1.0\linewidth]{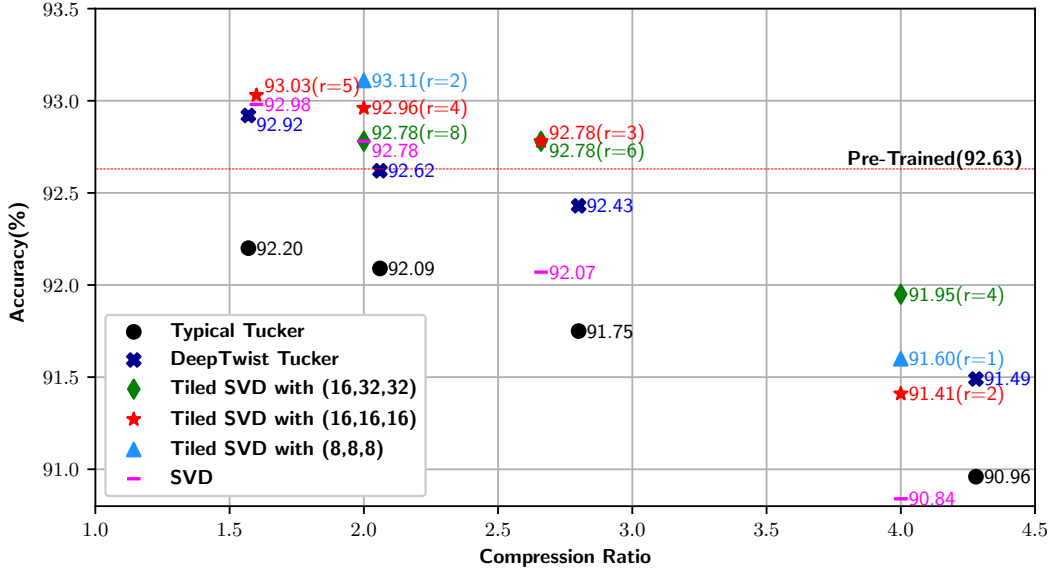}
\caption{Test accuracy of ResNet-32 model using CIFAR-10 with various target compression ratio and decomposition methods. Except the first small convolution layer, all layers are compressed by the same compression ratio. Convolution layers can be grouped according to 3 different $S$ values (=16, 32, or 64). For tiled SVD, three groups (of different $S$) are tiled in ($k_1 \times k_1$), ($k_2 \times k_2$), or ($k_3 \times k_3)$ tile size. ($k_1, k_2, k_3$) configuration is described in legends.  } 
\label{fig:svd}
\end{center}
\end{figure}

\section{Experimental Results}

In this section, we apply low-rank approximation trained by DeepTwist to various CNN models.

Figure~\ref{fig:svd} summarizes the test accuracy values of ResNet-32 (with CIFAR-10 dataset) compressed by various low-rank approximation techniques.
Note that tiled SVD and normal SVD are enabled only by DeepTwist, which obviates model structure modification during training.
All configurations in Figure~\ref{fig:svd} use the same learning rate scheduling and the number of training epochs as described in Section 3.
Results show that tiled SVD yields the best test accuracy and test accuracy is not highly sensitive to tile configuration.
SVD presents competitive model accuracy for small compression ratios.
As compression ratio increases, however, model accuracy using SVD significantly degrades.
From Figure~\ref{fig:svd}, tiled SVD associated with DeepTwist is clearly the best low-rank approximation scheme.


\begin{table}[t]
\begin{center}
\caption{Comparison on various low-rank approximation schemes of VGG19 (using CIFAR-10 dataset). To focus on convolution layers only, fully-connected layers are compressed by 8$\times$ and trained by DeepTwist. Then, fully-connected layers are frozen and convolution layers are compressed (except small layers of $S$<128) by Tucker decomposition or tiled SVD.} 
\label{table:vgg}
\begin{tabular}{cccccc}
\Xhline{2\arrayrulewidth}
\multicolumn{1}{c}{Comp. Scheme}  & Parameter &  Weight Size & FLOPs & Accuracy(\%) \\
\Xhline{2\arrayrulewidth}
 \mr{1}{Pre-Trained} & - & \mr{1}{18.98M} & \mr{1}{647.87M} & \mr{1}{92.37} \\
\hline
\mr{4}{Tucker\\ Decomposition \\ (Conventional\\Training)}  & $R_c$=0.6~~  & 9.14M (2.08$\times$) & 319.99M (2.02$\times$) & 91.97 \\ 
                                & $R_c$=0.5~~  & 6.71M (2.83$\times$) & 235.74M (2.75$\times$) & 91.79 \\
                                & $R_c$=0.45 & 5.49M (3.45$\times$) & 191.77M (3.38$\times$) & 91.36 \\
                                & $R_c$=0.4~~  & 4.61M (4.11$\times$) & 161.60M (4.01$\times$)& 91.11 \\
\hline
\mr{5}{Tiled\\SVD\\(DeepTwist)}  & 64$\times$64 ($r$=16) & 9.49M (2.00$\times$) & 316.28M (2.04$\times$) & 92.42 \\
                    & 64$\times$64 ($r$=11) & 6.52M (2.91$\times$) & 214.25M (3.02$\times$) & 92.33 \\
                    & 64$\times$64 ($r$=10) & 5.93M (3.20$\times$) & 193.85M (3.34$\times$) & 92.23\\
                    & 64$\times$64 ($r$=9)~~ & 5.55M (3.41$\times$) & 173.44M (3.73$\times$) & 92.22 \\
                    & 64$\times$64 ($r$=8)~~ & 4.74M (4.00$\times$) & 153.04M (4.33$\times$) & 92.07 \\
\Xhline{2\arrayrulewidth}                                                
\end{tabular}
\end{center}
\end{table}

We compare Tucker decomposition trained by a typical fine-tuning process and tiled SVD trained by DeepTwist using the VGG19 model\footnote{https://github.com/chengyangfu/pytorch-vgg-cifar10} with CIFAR-10.
Since this work mainly discusses compression on convolution layers, fully-connected layers of VGG19 are compressed and fixed before compression of convolution layers (refer to Appendix for details on the structure of VGG19).
Except for small layers with $S<128$ (that presents small compression ratio as well), all convolution layers are compressed with the same compression ratio.
During 300 epochs to train convolution layers, learning rate is initially 0.01 and is then halved every 50 epochs.
In the case of tiled SVD, $S_D$ is 300 and tile size is fixed to be 64$\times$64 (recall that the choice of $S_D$ and tile size do not affect model accuracy significantly).
As described in Table~\ref{table:vgg}, while Tucker decomposition with conventional fine-tuning shows degraded model accuracy through various $R_c$, DeepTwist-assisted tiled SVD presents noticeably higher model accuracy.



\begin{figure}[t]
\begin{center}
    \centering
    \includegraphics[width=1.0\linewidth]{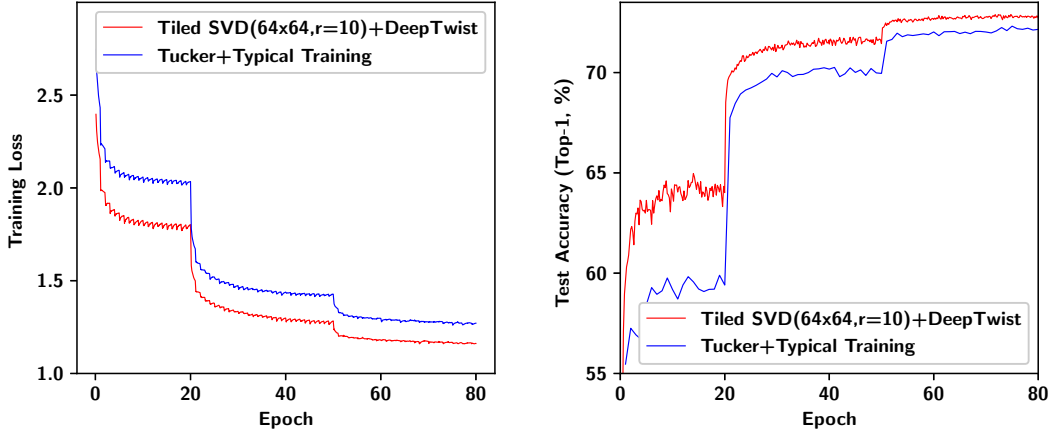}
\caption{Comparison of two compression methods on training loss and (top-1) test accuracy of ResNet-34 model using ImageNet. $S_D$=500.} 
\label{fig:imagenet}
\end{center}
\end{figure}

We also test our proposed low-rank approximation training technique with the ResNet-34 model\footnote{https://pytorch.org/docs/stable/torchvision/models.html} \cite{resnet} using the ImageNet dataset \cite{imagenet}.
A pre-trained ResNet-34 is fine-tuned for Tucker decomposition (with conventional training) or tiled SVD (with DeepTwist) using the learning rate of 0.01 for the first 20 epochs, 0.001 for the next 30 epochs, and 0.0001 for the remaining 30 epochs.
Similar to our previous experiments, the same compression ratio is applied to all layers except the layers with $S$<128 (such exceptional layers consist of 1.4\% of the entire model. Refer to Appendix for more details).
In the case of Tucker decomposition, selected convolution layers are compressed with $R_c=0.46$ to achieve an overall compression of $3.1\times$.
For tiled SVD, lowered matrices are tiled and each tile of (64$\times$64) size is decomposed with $r$=10 to match an overall compression of $3.1\times$.
As shown in Figure~\ref{fig:imagenet}, DeepTwist-based tiled SVD yields better training loss and test accuracy compared to Tucker decomposition with typical training.
At the end of the training epoch in Figure~\ref{fig:imagenet}, tiled SVD and Tucker decomposition achieves 73.00\% and 72.31\% for top-1 test accuracy, and 91.12\% and 90.73\% for top-5 test accuracy, while the pre-trained model shows 73.26\% (top-1) and 91.24\% (top-5).


\section{Conclusion}

In this paper, we propose a new training technique, DeepTwist, to efficiently compress CNNs by low-rank approximation.
DeepTwist injects noise to weights in the form of low-rank approximation without modifying the model structure and such noise enables wide local minima search exploration.
Compared to typical fine-tuning, DeepTwist improves training loss and test accuracy during the entire training process.
DeepTwist enables 2D SVD based on lowering and we demonstrate that tiled SVD provides even better test accuracy compared to Tucker decomposition.

\small

\bibliography{neurips_2019_svd}

\end{document}


\maketitle

\appendix

\section{Lowering Technique}

\begin{figure}[h]
\begin{center}
    \centering
    \includegraphics[width=1.0\linewidth]{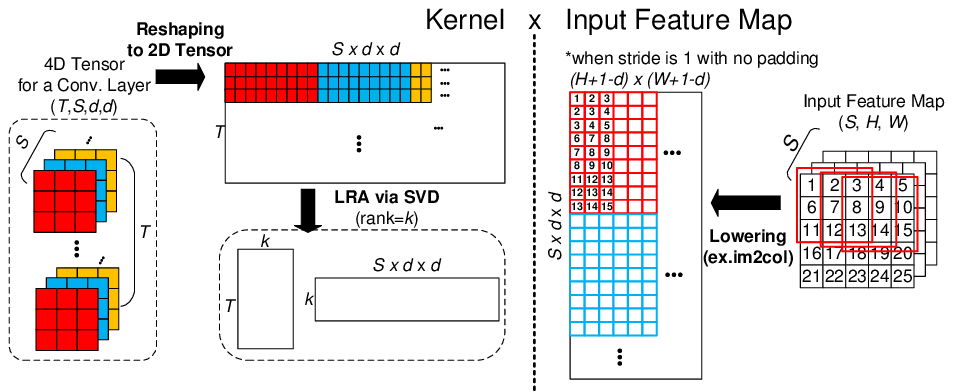}
\caption{An example of lowering technique using im2col.} 
\label{fig:lowering}
\end{center}
\end{figure}

Figure~\ref{fig:lowering} describes a kernel matrix reshaped from a 4D kernel tensor and an input feature map matrix in the form of a Toeplitz matrix.
At the cost of redundant memory usage to create a Toeplitz matrix, lowering enables matrix multiplication which can be efficiently implemented by BLAS libraries.
A kernal matrix can be decomposed by 2D SVD.

\section{Model Descriptions}

In this section, we describe model structures and layers selected for decomposition.
Small layers close to the input are not compressed because both weight size and compression rate are too small.

\begin{table}[h]\begin{center}
\caption{Convolution Layers of ResNet32 for CIFAR10} 
\label{table:app_resnet32}
\begin{tabular}{crrrr|c}
\Xhline{2\arrayrulewidth}
\# of layers & $T$ & $S$ & $d$ & Weight Size & Decomposed \\
\hline
1	& 16	& 3		& 3	& 0.4K (~~0.1\%) & No \\
10	& 16	& 16	& 3	& 22.5K (~~5.0\%) & Yes \\
1	& 32	& 16	& 3	& 4.5K (~~1.0\%) & Yes \\
9	& 32	& 32	& 3	& 81.0K (18.0\%) & Yes \\
1	& 64	& 32	& 3	& 18.0K (~~4.0\%) & Yes \\
9	& 64	& 64	& 3	& 324.0K (71.9\%) & Yes \\
\hline
\multicolumn{4}{c}{\textbf{Total}} & 450.4K (100.0\%)  \\
\Xhline{2\arrayrulewidth}                                                
\end{tabular}\end{center}\end{table}

\begin{table}[h]\begin{center}
\caption{Convolution and Fully-connected(FC) Layers of VGG19 for CIFAR10} 
\label{table:app_vgg19}
\begin{tabular}{c|crrrr|c}
\Xhline{2\arrayrulewidth}
 Type & \# of layers & $T$ & $S$ & $d$ & Weight Size & Decomposed \\
\hline

\mr{8}{Conv.} & 1 &	64 &	3 & 3 &	0.002M (~~0.01\%) & No \\
 & 1 &	64 &	64 & 3 & 0.035M (~~0.18\%) & No \\ 
 & 1 &	128 &	64 & 3 & 0.070M (~~0.36\%) & No \\
 & 1 &	128 &	128 & 3 & 0.141M (~~0.72\%) & Yes \\
 & 1 &	256 &	128 & 3 & 0.281M (~~1.44\%) & Yes \\
 & 3 &	256 &	256 & 3 & 1.688M (~~8.61\%) & Yes \\
 & 1 &	512 &	256 & 3 & 1.125M (~~5.74\%) & Yes \\
 & 7 &	512 &	512 & 3 & 15.75M (80.37\%) & Yes \\
\hline				
\mr{2}{FC} & 2 & 512 & 512 & - & 0.500M (~~2.55\%) & \mr{2}{Yes \\ (Pre-determined)} \\
& 1 & 512 & 10 & - & 0.005M (~~0.02\%) & \\
\hline			
\textbf{Total} & & & & & 19.597M (100.0\%)  \\
\Xhline{2\arrayrulewidth}                                                
\end{tabular}\end{center}\end{table}

\begin{table}[h]\begin{center}
\caption{Convolution Layers of ResNet34 for ImageNet} 
\label{table:app_resnet34}
\begin{tabular}{crrrr|c}
\Xhline{2\arrayrulewidth}
\# of layers & $T$ & $S$ & $d$ & Weight Size & Decomposed\\
\hline
1 & 64 & 3 & 7 & 0.01M (~~0.04\%) & No \\
6 & 64 & 64 & 3 & 0.21M (~~1.05\%) & No \\ 
1 & 128	& 64 & 3 & 0.07M (~~0.35\%) & No \\ 
7 & 128	& 128 & 3 & 0.98M (~~4.90\%) & Yes \\
1 & 256	& 128 & 3 & 0.28M (~~1.40\%) & Yes  \\
11 & 256 & 256 & 3 & 6.18M (30.77\%) & Yes\\
1 & 512	& 256 & 3 & 1.13M (~~5.59\%) & Yes \\
5 & 512	& 512 & 3 & 11.25M (55.94\%) & Yes\\
\hline
Total & & & & 20.11M (100.0\%) & \\
\Xhline{2\arrayrulewidth}                                                
\end{tabular}\end{center}\end{table}